%% file: main.tex
\documentclass[letterpaper,10pt,conference]{ieeeconf}

\IEEEoverridecommandlockouts
\usepackage{times}
\usepackage{mathptmx}
\usepackage{cite}
\usepackage{url}
\usepackage{amsmath}
\usepackage{amssymb}
\usepackage{graphicx}
\usepackage{booktabs}
\usepackage{multirow}
\usepackage{makecell}
\usepackage{balance}
\usepackage{xcolor}
\definecolor{projectlink}{HTML}{254B70}
\usepackage[hidelinks]{hyperref}

\makeatletter
\let\eventsync@template@makecaption\@makecaption
\long\def\@makecaption#1#2{%
  \ifx\@captype\@IEEEtablestring
    \eventsync@template@makecaption{#1}{#2}%
  \else
    \@IEEEfigurecaptionsepspace
    \setbox\@tempboxa\hbox{\footnotesize\mbox{#1.}\nobreak\hspace{0.5em}#2}%
    \ifdim\wd\@tempboxa>\hsize
      \parbox[t]{\hsize}{\footnotesize\noindent%
        \mbox{#1.}\nobreak\hspace{0.5em}#2}%
    \else
      \ifcenterfigcaptions
        \hbox to\hsize{\hfil\box\@tempboxa\hfil}%
      \else
        \hbox to\hsize{\box\@tempboxa\hfil}%
      \fi
    \fi
  \fi}
\makeatother

\title{\LARGE \bf
EventSync: Event-Driven Memory Updates\\
for History-Dependent Manipulation
}
\makeatletter
\renewcommand{\@IEEEauthorblockNstyle}{\normalfont\fontsize{11}{13}\selectfont}
\renewcommand{\@IEEEauthorblockAstyle}{\normalfont\fontsize{9}{11}\selectfont}
\renewcommand{\@IEEEauthorblockNinterlinespace}{13pt}
\renewcommand{\@IEEEauthorblockAinterlinespace}{11pt}
\renewcommand{\@IEEEauthorblockAtopspace}{0.7ex}
\makeatother

\author{%
\authorblockN{%
Jingjiao Zhao$^{1,2}$\hspace{0.9em} Dong Jing$^{1}$\hspace{0.9em} Mingyu Cao$^{2}$\hspace{0.9em} Ruochuan Shi$^{3,2}$\\[0.2ex]
Xiansheng Chen$^{2}$\hspace{0.9em} Pengwei Wang$^{2}$\hspace{0.9em} Zhiwu Lu$^{1,*}$%
\thanks{$^{*}$Corresponding author: Zhiwu Lu
(\href{mailto:luzhiwu@ruc.edu.cn}{\texttt{luzhiwu@ruc.edu.cn}}).}%
}
\authorblockA{%
$^{1}$Renmin University of China\qquad
$^{2}$Beijing Academy of Artificial Intelligence\\
$^{3}$Institute of Automation, Chinese Academy of Sciences\\[0.5ex]
{\urlstyle{same}\textcolor{projectlink}{\href{https://zjjJacinthe.github.io/eventsync-project}{\nolinkurl{zjjJacinthe.github.io/eventsync-project}}}}%
}
}

\begin{document}
\maketitle
\thispagestyle{empty}
\pagestyle{empty}

\input{sections/abstract}
\input{sections/introduction}
\input{sections/related_work}
\input{sections/method}
\input{sections/experiments}
\input{sections/conclusion}
\input{sections/acknowledgment}

\balance
\bibliographystyle{IEEEtran_compact}
\bibliography{references}

\end{document}

%% file: sections/abstract.tex
\begin{abstract}
History-dependent robotic manipulation requires task context and interaction history, as the current observation alone is insufficient to determine the next action.
Hierarchical frameworks use task memory for high-level subgoal reasoning and low-level policies for execution.
However, memory updates require accurate completion assessment: similar observations may correspond to different execution phases, and prematurely recording completion can mislead subsequent subgoal reasoning.
We present \textbf{EventSync}, a hierarchical framework that coordinates task reasoning and action execution through event-driven memory updates.
A vision-language model generates subgoals from task plans and interaction history, while a vision-language-action model executes them.
A lightweight \emph{Event Monitor} combines visual observations with end-effector trajectories to determine whether the current subgoal is complete.
The framework records the current subgoal in its completed-subgoal history and triggers subsequent reasoning only after obtaining completion evidence.
With visual history in the executor, EventSync achieves 64.5\% overall success on RoboMME, exceeding the best baseline reported in the benchmark by 20.0 percentage points.
Further analysis shows that this update mechanism helps build a reliable completed-subgoal history.
Experiments on four real-robot tasks with different memory requirements validate the framework's effectiveness in real-world robotic manipulation.
\end{abstract}

%% file: sections/introduction.tex
\section{Introduction}
\label{sec:introduction}

Real-world robotic manipulation often requires decisions based on previous interactions.
In assembly, tidying, and everyday service tasks, earlier requirements and interaction outcomes can determine what the robot should do next~\cite{kaelbling2013belief}.
History may also be needed to interpret the current phase of an ongoing operation.
Such tasks are non-Markovian with respect to instantaneous observations: robots must retain task context and interpret execution history to select subsequent goals and actions~\cite{kaelbling1998pomdp,memoryvla,sam2act,robomme}.

Prior work introduces memory into vision-language-action (VLA) models to use historical information~\cite{roboflamingo,memoryvla,eventvla,nativemem}, while other approaches assign task memory to a high-level reasoner that guides action execution through subgoals~\cite{memer,tau0vla}.
We argue that hierarchical frameworks are well suited to history-dependent manipulation.
The limited performance of hierarchical methods reported in existing evaluations~\cite{robomme} motivates us to revisit memory updates within this framework.
As illustrated in Fig.~\ref{fig:motivation}, an instruction issued by the high-level reasoner does not imply that the corresponding operation has been completed, and similar current poses may correspond to different execution phases.
If an unfinished operation is recorded as completed, subsequent reasoning may rely on an incorrect estimate of task progress and advance to the next subgoal prematurely.
For task-memory updates, completion judgments must be accurate enough to avoid premature goal transitions and timely enough to keep subsequent reasoning and actions aligned with execution progress.

\input{figures/motivation_pump}

We propose \textbf{EventSync}, a hierarchical framework that coordinates task reasoning and action execution through event-driven memory updates.
A \emph{Memory Reasoner} built on a vision-language model (VLM) obtains a task plan and combines it with interaction history to generate subgoals, which a VLA-based \emph{Action Executor} executes.
EventSync maintains an event state for the current operation, and a lightweight \emph{Event Monitor} with \textbf{87.39M} total parameters estimates completion from visual observations and end-effector trajectories.
The framework adds the current subgoal to the completed-subgoal history and triggers high-level reasoning for the next subgoal only after obtaining completion evidence.

With visual history in the executor, EventSync achieves a \textbf{64.5\%} overall success rate on RoboMME, exceeding FrameSamp+Modul, the best baseline reported in the benchmark, by \textbf{20.0 percentage points}~\cite{robomme}.
Our variant without visual history in the executor achieves \textbf{94.0\%} on SwingXtimes, compared with the \textbf{7.3\%} reported for the hierarchical framework GroundSG + QwenVL~\cite{robomme}.
Analysis of execution records further shows that updates guided by completion evidence help reduce premature progression and build a reliable completed-subgoal history.
Experiments on four real-robot tasks further demonstrate the effectiveness of combining task-memory reasoning with event-driven memory updates.

Our contributions are as follows:
\begin{itemize}
    \item \textbf{A hierarchical memory framework for history-dependent manipulation.}
    We introduce EventSync, in which a high-level reasoner manages task memory and generates subgoals, a low-level policy executes actions, and an independent lightweight monitor provides completion feedback.
    \item \textbf{Event-driven task-memory updates.}
    We design a lightweight Event Monitor that combines visual observations with end-effector trajectories to assess operation completion, using the resulting evidence to trigger updates to the completed-subgoal history and subsequent high-level reasoning.
    \item \textbf{Empirical evaluation in simulation and on real robots.}
    We evaluate EventSync on the RoboMME simulation benchmark and real-robot tasks to assess its effectiveness in history-dependent manipulation, with comparisons and ablations examining the role of event-driven memory updates.
\end{itemize}

%% file: figures/motivation_pump.tex
\begin{figure}[t]
\centering
\includegraphics[width=\columnwidth]{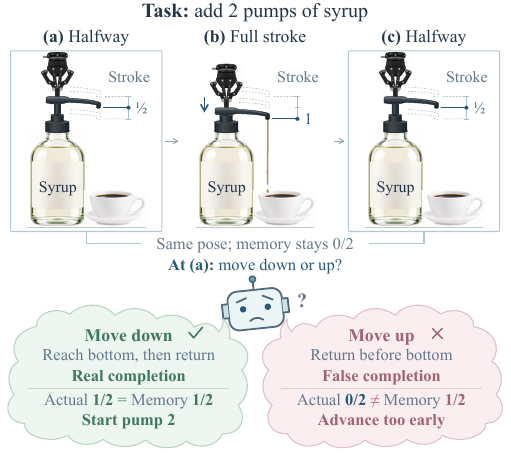}
\caption{Execution-phase ambiguity and erroneous memory updates. The pump head occupies the same position during pressing (a) and release (c), but must move downward in (a) and upward in (c). From (a), continuing to the bottom and returning to the fully released position completes the first pump. Recording a return before reaching the bottom as completion incorrectly triggers the second pump, causing the progress recorded in memory to exceed actual execution progress.}
\label{fig:motivation}
\end{figure}

%% file: sections/related_work.tex
\section{Related Work}
\label{sec:related_work}

\subsection{Memory for Robotic Manipulation}

Research on robot memory studies how to retain task-relevant history for subsequent decisions.
Existing methods preserve interaction history through recurrent states~\cite{roboflamingo}, spatial memory~\cite{sam2act}, and historical feature aggregation and retrieval~\cite{hamlet,memoryvla}.
Compressed context~\cite{contextvla,nativemem}, motion trajectories overlaid on images~\cite{tracevla}, and visual evidence selected by event predictions~\cite{eventvla} further provide temporal information for action prediction.
MemER~\cite{memer} retrieves historical keyframes for subgoal generation and runs reasoning asynchronously with control.
RoboMME~\cite{robomme} systematically evaluates how memory representations and their integration with action policies affect history-dependent manipulation.

\subsection{Vision-Language Reasoning and Closed-Loop Control}

Language-conditioned manipulation policies combine semantic and spatial representations to map instructions to actions~\cite{cliport,peract}.
Generalist robot policies and VLA models, including RT-1~\cite{rt1}, RT-2~\cite{rt2}, OpenVLA~\cite{openvla}, Octo~\cite{octo}, RDT-1B~\cite{rdt1b}, and the $\pi$ family~\cite{pi0,pi05}, support action generation from visual observations and language instructions.
High-level reasoning further organizes low-level execution through environment-grounded task planning~\cite{saycan,palme}, language-based action abstractions~\cite{rth,hirobot}, or predicted trajectories~\cite{hamster}.
Programmatic control~\cite{codeaspolicies} and 3D value maps~\cite{voxposer} also provide interfaces between language requirements and robot control.

Execution feedback allows subsequent decisions to adapt to actual observations.
Prior work supports replanning and execution recovery through success detection and scene feedback~\cite{inner_monologue}, precondition checking~\cite{progprompt}, constraint monitoring~\cite{doremi}, and failure explanation~\cite{reflect}.
Recently, $\tau_0$-VLA~\cite{tau0vla} has combined world-model-guided subtask search with a high-level model that jointly updates execution memory and generates subgoals.
EventSync separates completion monitoring from high-level goal reasoning.
A lightweight \emph{Event Monitor} combines visual observations and end-effector trajectories to provide completion evidence, which governs additions to the completed-subgoal history and triggers subsequent high-level reasoning.

%% file: sections/method.tex
\section{Method}
\label{sec:method}

\subsection{Problem Formulation and Framework Overview}
\label{sec:method_overview}

We consider history-dependent manipulation specified by a language instruction $\ell$, with a demonstration video $D$ provided for some tasks.
At control step $t$, the robot receives an observation $o_t$ containing visual and proprioceptive information.
The current image in this observation is denoted by $I_t$.
The system uses task context and interaction history to select the current subgoal and generate corresponding actions.

As shown in Fig.~\ref{fig:eventsync-framework}, EventSync coordinates task reasoning and action execution through event-driven memory updates.
We represent task state using a history of completed subgoals and the current subgoal.
At the start of a task, the VLM-based Memory Reasoner generates the first subgoal from task context and the initial observation, and the VLA-based Action Executor executes it.

During execution, the Event Monitor uses visual observations and end-effector trajectories to estimate event progress associated with the current subgoal.
Upon receiving completion evidence, the system adds the current subgoal to the completion history and triggers the Memory Reasoner to infer the next subgoal from the updated history.
The Action Executor then generates an action chunk conditioned on the new subgoal.

\input{figures/method}

Let $g_t$ denote the current subgoal used for execution at control step $t$.
The executor combines it with observation $o_t$ to generate an action chunk:
\begin{equation}
    a_{t:t+K_a-1}=\pi_\phi(o_t,g_t,\mathbf M_t^{\mathrm{exec}}),
    \label{eq:action_executor}
\end{equation}
where $K_a$ is the action-chunk length and $\mathbf M_t^{\mathrm{exec}}$ denotes optional visual-history features used by the executor.

\subsection{Task Memory and Subgoal Reasoning}
\label{sec:task_memory}

The Memory Reasoner serves two roles: initial planning and online subgoal reasoning.
At the start of a task, it plans from the language instruction.
When a demonstration video is provided, it organizes its understanding of the demonstration into a task plan that remains fixed for reuse during execution.
During execution, it combines current observations with historical information to infer the subgoal to execute.

For memory-dependent tasks with demonstration videos, understanding the demonstration requires identifying its events and their temporal relations~\cite{densecaption}.
To strengthen temporal understanding, we require the Memory Reasoner to produce a structured event sequence for $D$, explicitly describing event content, time intervals, and repeated occurrences.
Building on this understanding, the reasoner uses the requirements in instruction $\ell$ to form an initial task plan $P$ for subsequent execution.
The event sequence and task plan together form a structured demonstration summary, connecting demonstrated events with subsequent execution requirements.

The Memory Reasoner's task memory consists of textual memory and visual history.
Textual memory contains the language instruction $\ell$ and the history of completed subgoals $\mathcal G_t$.
For tasks with demonstration videos, it also includes the task plan $P$ from the demonstration summary.
The instruction and plan provide fixed task requirements, while the completion history is updated during interaction: a completed subgoal is added only after the Event Monitor provides completion evidence.
The reasoner can therefore select subsequent subgoals using both task requirements and completed operations.

Visual history provides evidence of interaction processes not explicitly recorded in textual memory, including changes in the environment and changes caused by robot actions.
We uniformly sample images from previous interactions and arrange them chronologically to form $\mathcal H_t$, which is provided to the Memory Reasoner alongside the current image $I_t$.
Let $g_t^{-}$ denote the subgoal before the update.
Combining textual memory, $g_t^{-}$, and visual evidence, the reasoner generates the next subgoal:
\begin{equation}
    g_t=R_\theta(\ell,P,\mathcal G_t,g_t^{-},I_t,\mathcal H_t),
    \label{eq:goal_reasoning}
\end{equation}
The subgoal $g_t$ contains an action instruction and spatial coordinates grounded in the current image, specifying the operation and its target location for the Action Executor.
Its action instruction also determines the event to monitor.

We train the Memory Reasoner with token cross-entropy on structured demonstration outputs and online subgoal targets.

\subsection{Event Monitoring}
\label{sec:event_monitoring}

The Event Monitor, detailed in the lower part of Fig.~\ref{fig:eventsync-framework}, provides evidence for task-progress updates in two settings: operations performed by the robot and environmental state changes.
These settings require determining whether the current operation has completed and whether a new target event has occurred between observations, respectively.
When a new subgoal begins, the system creates an event instance and maps its action description to an event type using predefined rules.
We reset the event's start time and monitoring state for each new instance, even when its action description matches a previous one.

For operations performed by the robot, end-effector trajectories provide motion cues that complement visual observations when assessing completion.
For example, when placing an object into a container, the end effector may pass through similar positions while lowering the object and withdrawing.
The trajectory supplies motion direction and progression.
Together with the images at the operation's start and the current step, this motion information helps distinguish operation phases and assess placement completion.
The monitor therefore fuses visual and trajectory tokens to estimate the completion probability.

For environmental state changes, event detection relies on how the event unfolds across observations.
For example, in conveyor picking, a robot may need to identify an object passing a picking location.
A single passage can span multiple frames, which the monitor must recognize as one event.
The monitor therefore retains temporal state across queries and predicts the number of new target events between successive queries, termed the event increment.
The system accumulates these increments to track event progress and uses this progress together with task requirements to determine when to trigger the subsequent operation.

We extract image tokens with a frozen DINOv2-Base visual encoder~\cite{dinov2}.
For event instance $k$, $\mathbf V_{k,t}$ collects visual tokens from the event's start, the current observation, and causal history, $\mathbf T_{k,t}$ denotes end-effector trajectory tokens, and $u_k$ specifies the event condition determined by the subgoal and task requirements.
The unified prediction is
\begin{equation}
\begin{aligned}
    &\left(p_t^c,\boldsymbol p_t^{\mathrm{inc}},\mathbf s_t\right) \\
    &\quad=F_\psi\!\left(\mathbf V_{k,t},\mathbf T_{k,t},u_k,\mathbf s_{t^-}\right),
\end{aligned}
\label{eq:event_prediction}
\end{equation}
where $p_t^c$ is the completion probability for an operation performed by the robot and $\boldsymbol p_t^{\mathrm{inc}}$ is the probability distribution over event increments.
The time $t^-$ denotes the preceding monitoring query.
The states $\mathbf s_{t^-}$ and $\mathbf s_t$ are the temporal state retained before the current query and its updated value, respectively.
The two predictions share the visual encoder and Transformer~\cite{transformer} fusion parameters.
Completion prediction uses the images at the operation's start and the current step together with the motion trajectory, while event-increment prediction additionally uses a GRU~\cite{gru} to retain information across queries.
We supervise these predictions with operation-completion labels and event-increment labels, respectively.

%% file: figures/method.tex
\begin{figure*}[t]
\centering
\includegraphics[width=\textwidth]{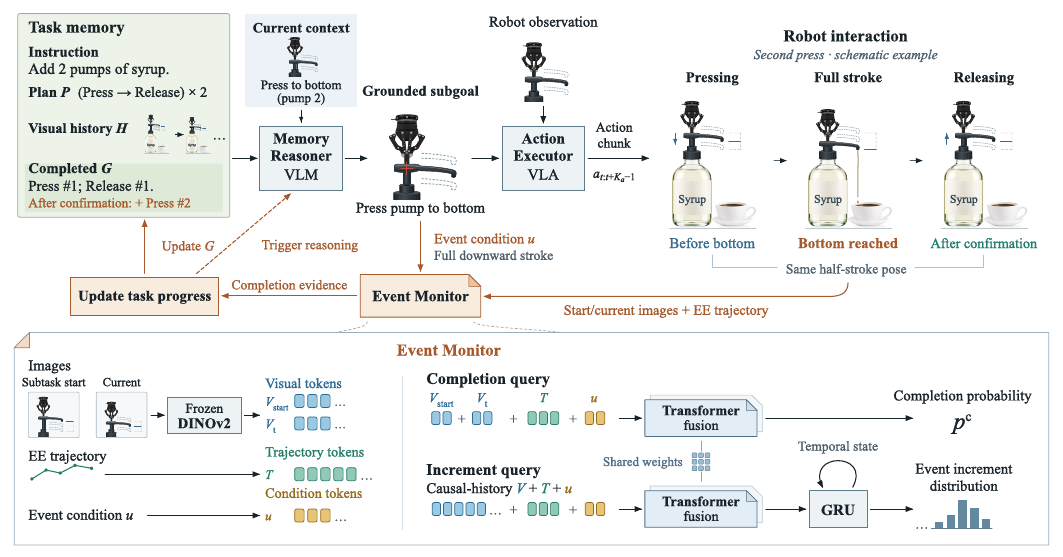}
\caption{Overview of EventSync. Top: the Memory Reasoner combines task memory and current context to generate grounded subgoals for the Action Executor. Completion evidence from the Event Monitor updates the completed-subgoal history and triggers subsequent reasoning. Bottom: visual, trajectory, and event-condition tokens feed shared Transformer fusion for completion prediction and recurrent event-increment prediction.}
\label{fig:eventsync-framework}
\end{figure*}

%% file: sections/experiments.tex
\section{Experiments}
\label{sec:experiments}

Our experiments evaluate EventSync on history-dependent manipulation and examine the role of event-driven memory updates.
We first compare overall performance on RoboMME, then analyze the effects of memory updates on subgoal progression, the design of the Event Monitor, and the contributions of individual memory components to task execution.
Real-robot experiments further assess the framework on tasks spanning four categories of memory requirements.

\subsection{Experimental Setup}
\label{sec:experimental_setup}

\textbf{Simulation tasks.}
We evaluate on the 16 history-dependent tasks in RoboMME~\cite{robomme}, covering four categories: Counting, Permanence, Reference, and Imitation.
Each task is evaluated over 50 episodes.

\textbf{Real-robot setup.}
As shown in Fig.~\ref{fig:real-robot-setup}, our platform consists of a Franka Panda arm with a Robotiq gripper, using 7-DoF joint-space control.
The executor receives current observations from the right external camera and the wrist camera, while the Memory Reasoner and Event Monitor use the right external camera.
The left camera is used only for recording.
The four real-robot tasks correspond to the Counting, Permanence, Reference, and Imitation categories.
\emph{SwingXTimes} requires a specified number of back-and-forth swings; \emph{UnmaskCube} requires uncovering the target cube after occlusion and cup swapping; \emph{ImagePickPlace} requires placing all cubes specified by a reference image into a basket in any order; and \emph{SequencePickPlace} requires placing cubes into a basket in the order shown in a demonstration video.
Both pick-and-place tasks end by ringing a bell.
We collect 90, 90, 150, and 150 trajectories for these tasks, respectively.
We train on a mixture of data from all four tasks.
Action chunk lengths are 20 during training and 16 during inference.

\input{figures/real_robot_setup}

\textbf{Baselines.}
On RoboMME, we compare against the base action policy $\pi_{0.5}$, FrameSamp+Modul, which incorporates visual history into the executor, and GroundSG + QwenVL and MemER, which use high-level reasoning.
We use the results reported in RoboMME, including its reproduction of MemER.
EventSync is evaluated with two executor configurations: one without visual history and one with FrameSamp+Modul visual history.
Comparisons of update mechanisms and memory components are described in their respective analysis subsections.

\textbf{Training annotations.}
For tasks with demonstration videos, we use GPT-5.4 to annotate demonstrations in both simulation and real-robot data.
The supervision combines structured event sequences describing operations, temporal intervals, and repetitions with initial task plans based on the language instructions.
These targets are used to fine-tune the Memory Reasoner.
The Event Monitor is trained using execution event labels.
In simulation, we use existing event annotations, with SwingXtimes completion boundaries relabeled from visual observations.
For real-robot data, we segment operation phases using end-effector states to generate completion labels.
We resample training examples near event boundaries to help the Event Monitor distinguish states before and after an event.

\textbf{Implementation.}
The Memory Reasoner is based on Qwen3-VL-4B-Instruct~\cite{qwen3vl} and fine-tuned for two epochs using LoRA~\cite{lora} with rank 16, scaling factor 32, and a learning rate of $10^{-4}$.
The two Action Executor configurations are fine-tuned separately from $\pi_{0.5}$~\cite{pi05} for 80,000 steps on simulation data and 20,000 steps on real-robot data.
The Event Monitor uses a two-layer Transformer to fuse 128-dimensional input features and contains 87.39M total parameters, including the visual encoder.
It is trained with AdamW~\cite{adamw} for 30 epochs at a learning rate of $3\times10^{-4}$.

For tasks with demonstration videos, the Memory Reasoner uniformly samples up to 64 frames for initial planning.
During online reasoning, it receives the current image and up to four causal history frames in temporal order.
The executor with visual history uniformly samples up to 32 frames from accumulated observations and compresses each frame into 16 visual tokens, yielding a memory input budget of 512 tokens.
The Event Monitor uses 64 visual tokens per image and 16 sampled points from the end-effector trajectory.
In simulation, it checks robot operations at action-chunk updates and environmental events every four simulator steps.
For real-robot execution, monitoring runs independently with a query interval set to 0.5~s.
In the asynchronous baseline, the simulator continues advancing during high-level inference and executes a hold action when a new action chunk is unavailable.
Training and evaluation use NVIDIA H100 GPUs.
All other simulation settings follow RoboMME unless otherwise specified.

\subsection{Results on RoboMME}
\label{sec:robomme_results}

Table~\ref{tab:robomme_main} summarizes task success rates on RoboMME.
EventSync achieves an overall success rate of 64.5\%, exceeding the 44.5\% reported for FrameSamp+Modul in the original benchmark paper~\cite{robomme} by 20.0 percentage points and demonstrating the framework's overall effectiveness.
With no visual history in the executor, EventSync still achieves 62.0\% overall success, outperforming GroundSG + QwenVL, which also uses a hierarchical architecture, by 29.3 percentage points.
In particular, on SwingXtimes, which requires continuous tracking of progress through repeated operations, this configuration achieves 94.0\% success, an improvement of 86.7 percentage points over GroundSG + QwenVL's 7.3\%.
These results demonstrate the potential of hierarchical frameworks for manipulation that requires tracking task progress.

\input{tables/robomme_main_results}

\subsection{Event-Driven Memory Updates}
\label{sec:memory_updates}

We evaluate the role of the Event Monitor using the base executor.
EventSync uses event evidence to trigger high-level reasoning and update the completed-subgoal history.
Without the Event Monitor, we consider fixed-schedule and asynchronous baselines.
The fixed schedules invoke high-level reasoning once every one, two, or four action-chunk requests, respectively, reusing the current subgoal for intermediate requests.
These configurations and EventSync use synchronous high-level calls.
The asynchronous configuration uses an optimized inference backend to overlap high-level reasoning with action execution, operating at approximately 1.5~Hz and 15~Hz, respectively.
We compare task success, online high-level calls, and mean episode wall-clock time across configurations, normalizing calls and time to Fixed 1:1.

Compared with Fixed 1:1, EventSync improves success by 6.5 percentage points while reducing high-level calls by 70.8\%.
As shown in Fig.~\ref{fig:eventsync-update-tradeoff}(b), mean episode wall-clock time decreases from 78.09~s to 46.74~s, a reduction of 40.1\%.
With similar high-level call counts, EventSync achieves 62.0\% success versus 52.1\% for Fixed 4:1 in Fig.~\ref{fig:eventsync-update-tradeoff}(a), a gain of 9.9 percentage points.
This comparison supports using execution progress to determine when to update memory and subgoals.
Fixed 4:1 and optimized asynchronous execution are faster, but achieve lower success rates of 52.1\% and 53.0\%, respectively.

\input{figures/eventsync_update_tradeoff}

We compare SwingXtimes execution records using simulator-provided subgoal completion states to track physical progress in task order.
A completion record is premature if written before actual completion; only the first record of each nonterminal operation is counted.
As shown in Fig.~\ref{fig:eventsync-swing-timeline}, EventSync reduces episodes with premature records from 64.0\% (32/50) to 10.0\% (5/50) relative to Fixed 1:1, and the proportion of premature records from 42.5\% (134/315) to 5.7\% (18/314).
The episode-averaged fraction of environment steps with memory ahead falls from 27.7\% to 3.5\%.

In the paired example in Fig.~\ref{fig:eventsync-swing-timeline}, Fixed 1:1 switches to B while the cube is returning to A, recording six arrivals despite only two simulator-verified arrivals in the required order.
EventSync waits for completion evidence before advancing the history and completes all three required rounds.

\input{figures/eventsync_swing_timeline}

\subsection{Event Monitor Analysis}
\label{sec:event_monitor_analysis}

We analyze the Event Monitor's trajectory input and monitoring model while keeping the Memory Reasoner, the corresponding executor, and memory update rules unchanged.
For the trajectory ablation, we remove the end-effector trajectory input and retrain the monitor.

In addition, we replace the Event Monitor with a fine-tuned VLM to examine how the monitoring model affects task execution.
The baseline uses a separately LoRA-fine-tuned Qwen3-VL-4B with the same subgoal conditions and completion labels as the Event Monitor.
Given the subgoal condition, the subgoal's start frame, and the current frame, the model predicts ``complete'' or ``incomplete''.
To convey motion, the end-effector trajectory is overlaid on the start frame as connected blue lines, with lighter segments indicating earlier motion and darker segments indicating more recent motion.

\input{tables/completion_monitor_comparison}

As shown in Table~\ref{tab:completion_monitor_comparison}, the Event Monitor achieves higher task success than the fine-tuned VLM completion monitor with both executors under the same high-level reasoning and memory update rules, while using a smaller model and providing lower response latency.
It contains 87.39M total parameters including the visual encoder, approximately 2.2\% of the 4B VLM's parameter count.
Its mean latency per monitoring query is 19.52~ms, compared with 215.84~ms for the fine-tuned VLM.
These results show that a lightweight dedicated monitor can efficiently provide completion evidence to support high-level task-memory updates.

Removing trajectory inputs reduces success by 3.0 and 2.0 percentage points with the base and FrameSamp+Modul executors, respectively, supporting the use of motion cues in completion monitoring for memory updates.
Together with the preceding analysis of premature progression, these results support using completion evidence from execution to maintain the completed-subgoal history for subsequent subgoal reasoning.

\subsection{Analysis of Memory Components}
\label{sec:memory_components}

We examine how task memory supports subgoal reasoning within EventSync and how it combines with executor memory.
As summarized in Table~\ref{tab:memory_components}, all configurations retain the Event Monitor, allowing us to analyze memory components under the same event-driven update mechanism.

\input{tables/memory_components}

\textbf{Structured demonstration understanding.}
Initial planning constructs reusable task memory for subsequent subgoal reasoning.
We compare two training objectives: predicting only the task plan, and jointly predicting a structured event sequence and the task plan.
Both configurations use up to 64 demonstration frames.
As shown in Table~\ref{tab:memory_components}, joint prediction achieves higher overall success with both executors.
On the nine tasks with demonstration videos, success improves from 49.3\% to 55.8\% with the base executor and from 55.3\% to 59.6\% with FrameSamp+Modul.
These results show that training the reasoner to jointly predict structured event sequences and task plans improves subsequent execution that relies on demonstration memory.

\textbf{High-level online visual history.}
Online visual history provides observations of environmental and robot-state changes alongside the task plan and completed-subgoal history.
We remove history frames from the Memory Reasoner while retaining the current image, all textual memory, and the Event Monitor inputs.
As shown in Table~\ref{tab:memory_components}, success drops from 62.0\% to 49.9\% with the base executor and from 64.5\% to 54.5\% with FrameSamp+Modul, decreases of 12.1 and 10.0 percentage points, respectively.
These results show that combining event-driven progress updates with visual evidence from the interaction benefits subsequent subgoal reasoning.

\textbf{Executor visual history.}
With high-level task memory and event-driven updates retained, adding FrameSamp+Modul improves overall success from 62.0\% to 64.5\%.
This result supports combining EventSync's hierarchical coordination with executor memory: the high-level reasoner uses task memory to select subgoals, while the executor also benefits from historical observations when generating actions.

\subsection{Real-Robot Experiments}
\label{sec:real_robot_results}

We evaluate the four physical tasks shown in Fig.~\ref{fig:real-robot-setup}, comparing $\pi_{0.5}$~\cite{pi05}, FrameSamp+Modul~\cite{robomme}, EventSync without the Event Monitor, and EventSync.
Each method is tested in 10 trials per task, with success determined by the physical task outcome.
SwingXTimes includes instructions to complete one, two, or three back-and-forth cycles.
UnmaskCube covers zero, one, and two cup swaps, including swap sequences absent from the training set.
The reference image in ImagePickPlace specifies one or two target cubes.
SequencePickPlace requires placing cubes into the basket in the order shown in a demonstration video.
The videos also contain cubes being picked up without being placed in the basket, requiring the robot to identify the actual placement sequence.
All methods use identical test conditions.
For tasks requiring demonstration videos, the videos are recorded in advance and shared across methods.
Table~\ref{tab:real_robot_results} reports successful trials and total trials.

\input{tables/real_robot_results}

EventSync obtains the highest success count on all four tasks.
Without the Event Monitor, the hierarchical configuration already achieves 8/10 on ImagePickPlace and 7/10 on SequencePickPlace, compared with 4/10 and 2/10 for FrameSamp+Modul.
These results show that a hierarchical framework can perform these tasks more effectively by interpreting reference information and generating subgoals at the high level.

The effect of event monitoring is most pronounced on SwingXTimes, where adding the monitor increases success from 0/10 to 9/10; $\pi_{0.5}$ and FrameSamp+Modul achieve 1/10 and 5/10, respectively.
Both hierarchical configurations achieve 7/10 on UnmaskCube and SequencePickPlace, while ImagePickPlace improves from 8/10 to 9/10 with the monitor.
These results support coupling task-memory reasoning with completion-based progress updates in physical manipulation, particularly when tracking repeated operations.

%% file: figures/real_robot_setup.tex
\begin{figure*}[t]
\centering
\includegraphics[width=\textwidth]{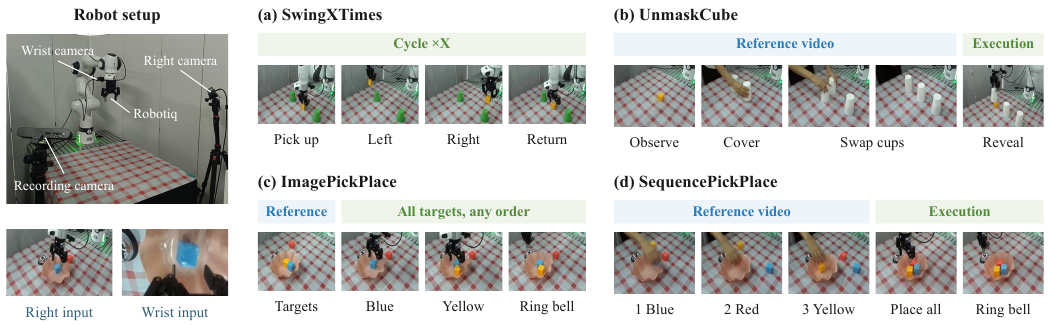}
\caption{Real-robot setup and four history-dependent tasks: (a) SwingXTimes, (b) UnmaskCube, (c) ImagePickPlace, and (d) SequencePickPlace. Blue headers indicate reference inputs; green headers indicate execution.}
\label{fig:real-robot-setup}
\end{figure*}

%% file: tables/robomme_main_results.tex
\begin{table*}[t]
\caption{Task success rates (\%) on RoboMME. Each category includes four task results and their average; Overall AVG is the average over all 16 tasks. Results for the four baselines are means over repeated evaluations reported in RoboMME~\cite{robomme}. Bold indicates the highest value in each column. Results are displayed to one decimal place.}
\label{tab:robomme_main}
\centering
\begingroup
\setlength{\tabcolsep}{2pt}
\renewcommand{\arraystretch}{1.2}
\scriptsize
\resizebox{\textwidth}{!}{%
\begin{tabular}{@{}l*{21}{c}@{}}
\toprule
\multirow{3}{*}{Method} & \multicolumn{5}{c}{Counting} & \multicolumn{5}{c}{Permanence} & \multicolumn{5}{c}{Reference} & \multicolumn{5}{c}{Imitation} & \multirow{3}{*}{\makecell{Overall\\AVG}} \\
\cmidrule(lr){2-6}\cmidrule(lr){7-11}\cmidrule(lr){12-16}\cmidrule(lr){17-21}
& Bin & Pick & Swing & Stop & \multirow{2}{*}{AVG}
& Video & Button & Video & Button & \multirow{2}{*}{AVG}
& Pick & Video & Video & Video & \multirow{2}{*}{AVG}
& Move & Insert & Pattern & Route & \multirow{2}{*}{AVG} & \\
& Fill & Xtimes & Xtimes & Cube &
& Umsk & Umsk & UmskS & UmskS &
& HighL & Repick & PlcBtn & PlcOrd &
& Cube & Peg & Lock & Stick & & \\
\midrule
$\pi_{0.5}$ & 30.0 & 42.9 & 35.6 & 6.7 & 28.8 & 20.4 & 22.2 & 18.7 & 6.7 & 17.0 & 11.3 & 0.4 & 31.1 & 25.8 & 17.2 & 26.0 & 1.6 & 2.9 & 4.7 & 8.8 & 17.9 \\
GroundSG + QwenVL & 52.0 & 92.7 & 7.3 & 0.0 & 38.0 & 88.7 & 24.0 & 30.7 & 14.0 & 39.3 & 15.1 & 25.3 & 54.0 & 31.8 & 31.6 & 71.6 & 3.3 & 6.7 & 6.0 & 21.9 & 32.7 \\
MemER & 56.7 & 79.3 & 59.3 & 0.0 & 48.8 & 81.3 & 72.0 & 38.0 & 21.3 & 53.2 & \textbf{70.7} & 25.3 & 30.0 & 26.0 & 38.0 & 82.7 & 6.7 & 16.7 & 12.0 & 29.5 & 42.4 \\
FrameSamp+Modul & 39.6 & 87.3 & 92.0 & \textbf{42.0} & 65.2 & 32.7 & 25.1 & 24.4 & 18.2 & 25.1 & 22.9 & 30.4 & 60.0 & 32.0 & 36.3 & 77.8 & 7.6 & \textbf{53.6} & \textbf{66.7} & \textbf{51.4} & 44.5 \\
\midrule
EventSync & 76.0 & \textbf{100.0} & 94.0 & 40.0 & \textbf{77.5} & 96.0 & 82.0 & 56.0 & \textbf{32.0} & 66.5 & 66.0 & 42.0 & \textbf{84.0} & 70.0 & 65.5 & 84.0 & 14.0 & 40.0 & 16.0 & 38.5 & 62.0 \\
\addlinespace[1pt]
\makecell[cl]{EventSync\\(w FrameSamp+Modul)} & \textbf{78.0} & \textbf{100.0} & \textbf{96.0} & 30.0 & 76.0 & \textbf{98.0} & \textbf{100.0} & \textbf{60.0} & 24.0 & \textbf{70.5} & 68.0 & \textbf{44.0} & \textbf{84.0} & \textbf{72.0} & \textbf{67.0} & \textbf{86.0} & \textbf{26.0} & 40.0 & 26.0 & 44.5 & \textbf{64.5} \\
\bottomrule
\end{tabular}%
}
\endgroup
\end{table*}

%% file: figures/eventsync_update_tradeoff.tex
\begin{figure}[t]
\centering
\includegraphics[width=\columnwidth]{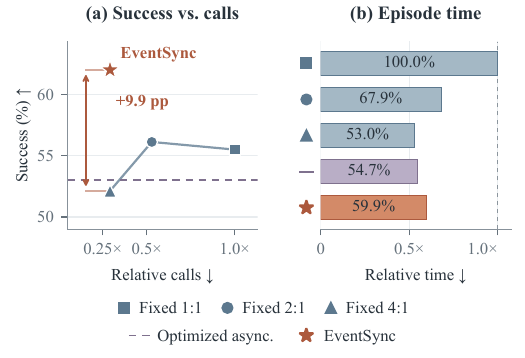}
\caption{Task-memory and subgoal update strategies with the base executor. Only EventSync uses the Event Monitor. (a) Success versus online high-level calls; the dashed line shows asynchronous success. (b) Mean episode wall-clock time. Calls and time are normalized to Fixed 1:1.}
\label{fig:eventsync-update-tradeoff}
\end{figure}

%% file: figures/eventsync_swing_timeline.tex
\begin{figure*}[t]
\centering
\includegraphics[width=\textwidth]{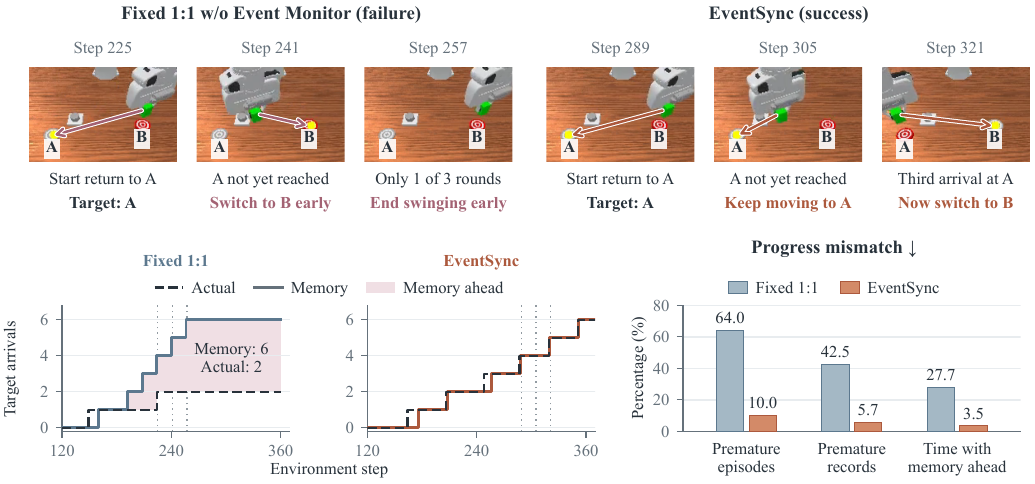}
\caption{Memory--execution alignment on SwingXtimes. Top: paired executions; arrows indicate commanded targets. Bottom left: cumulative arrivals recorded in memory (solid) and verified in task order by the simulator (dashed); shading marks memory ahead, and dotted lines mark displayed frames. Bottom right: percentages of premature episodes, premature completion records (one per nonterminal operation), and steps with memory ahead (episode average). Statistics cover 50 paired episodes per method.}
\label{fig:eventsync-swing-timeline}
\end{figure*}

%% file: tables/completion_monitor_comparison.tex
\begin{table}[t]
\caption{Completion monitor comparison on RoboMME. Overall task success rates (\%) are displayed to one decimal place. Bold indicates the highest value in each column.}
\label{tab:completion_monitor_comparison}
\centering
\begingroup
\footnotesize
\setlength{\tabcolsep}{3pt}
\renewcommand{\arraystretch}{1.2}
\begin{tabular*}{\columnwidth}{@{\extracolsep{\fill}}lcc@{}}
\toprule
Completion monitor & \makecell{Base\\executor} & \makecell{FrameSamp+Modul\\executor} \\
\midrule
Qwen3-VL-4B & 59.8 & 61.4 \\
\makecell[cl]{Event Monitor\\(w/o trajectory)} & 59.0 & 62.5 \\
\addlinespace[1pt]
Event Monitor & \textbf{62.0} & \textbf{64.5} \\
\bottomrule
\end{tabular*}
\endgroup
\end{table}

%% file: tables/memory_components.tex
\begin{table}[t]
\caption{Memory component ablations on RoboMME. $\checkmark$ and $\times$ indicate retained and removed components. Values are overall task success rates (\%) with the two executors.}
\label{tab:memory_components}
\centering
\begingroup
\footnotesize
\setlength{\tabcolsep}{3pt}
\renewcommand{\arraystretch}{1.2}
\begin{tabular*}{\columnwidth}{@{\extracolsep{\fill}}cccc@{}}
\toprule
\makecell{Structured\\event prediction} & \makecell{High-level\\visual history} & \makecell{Base\\executor} & \makecell{FrameSamp+Modul\\executor} \\
\midrule
$\checkmark$ & $\checkmark$ & \textbf{62.0} & \textbf{64.5} \\
$\times$ & $\checkmark$ & 57.1 & 61.5 \\
$\checkmark$ & $\times$ & 49.9 & 54.5 \\
\bottomrule
\end{tabular*}
\endgroup
\end{table}

%% file: tables/real_robot_results.tex
\begin{table}[t]
\caption{Real-robot task success (successful trials / total trials). Bold indicates the highest count in each task.}
\label{tab:real_robot_results}
\centering
\begingroup
\footnotesize
\setlength{\tabcolsep}{3pt}
\renewcommand{\arraystretch}{1.2}
\begin{tabular*}{\columnwidth}{@{\extracolsep{\fill}}lcccc@{}}
\toprule
Method & \makecell{Swing\\XTimes} & \makecell{Unmask\\Cube} & \makecell{Image\\PickPlace} & \makecell{Sequence\\PickPlace} \\
\midrule
$\pi_{0.5}$ & 1/10 & 3/10 & 2/10 & 2/10 \\
FrameSamp+Modul & 5/10 & 5/10 & 4/10 & 2/10 \\
\midrule
\makecell[cl]{EventSync\\(w/o Event Monitor)} & 0/10 & \textbf{7/10} & 8/10 & \textbf{7/10} \\
\addlinespace[1pt]
EventSync & \textbf{9/10} & \textbf{7/10} & \textbf{9/10} & \textbf{7/10} \\
\bottomrule
\end{tabular*}
\endgroup
\end{table}

%% file: sections/conclusion.tex
\section{Conclusion}
\label{sec:conclusion}

We presented EventSync, a hierarchical framework in which the Memory Reasoner generates subgoals from task plans and interaction history for the Action Executor.
A lightweight Event Monitor provides completion evidence to determine when to update the completed-subgoal history and invoke subsequent reasoning.
Comparisons and execution-record analysis on RoboMME suggest that reducing premature subgoal progression through event-driven updates contributes to higher task success.
Experiments on four real-robot tasks further demonstrate the framework's ability to use reference information and track repeated operations.
Together, these results show that event monitoring strengthens the hierarchical framework, enabling EventSync to handle history-dependent manipulation effectively.

%% file: sections/acknowledgment.tex
\section*{ACKNOWLEDGMENT}

We used GPT-Image-2 to generate the syrup bottle, flowing liquid, and coffee cup icons in the motivation and method illustrations shown in Figs.~\ref{fig:motivation} and~\ref{fig:eventsync-framework}.